\documentclass[letterpaper, 10 pt, conference]{ieeeconf}  

\usepackage{cite}
\usepackage{amsmath,amssymb,amsfonts}
\usepackage{graphicx} 
\usepackage{xcolor}
\usepackage{subcaption}
\PassOptionsToPackage{hyphens}{url}
\usepackage{comment}

\makeatletter
\let\NAT@parse\undefined
\makeatother
\usepackage{hyperref}

\usepackage{caption}
\usepackage{stfloats}
\usepackage{booktabs}

\usepackage{tikz}
\usetikzlibrary{shapes.geometric, arrows, positioning}

\definecolor{darkgreen}{RGB}{0,164,0}

\tikzstyle{decision} = [diamond, draw, fill=blue!20, 
text width=4.5em, text badly centered, inner sep=0pt, font=\small]
\tikzstyle{process} = [rectangle, draw, fill=orange!30, 
text width=5em, text centered, rounded corners, minimum height=4em, font=\small]
\tikzstyle{line} = [draw, -latex']

\title{\LARGE \bf
Low-Latency Quasi-Static Modeling of UAV Tether Aerodynamics
}

\author{Max Beffert and Andreas Zell\\
Cognitive Systems Group, University of Tübingen, Germany\\
\texttt{max.beffert@uni-tuebingen.de}
}

\newcommand\copyrighttext{%
	\footnotesize \textcopyright 2026 IEEE. Personal use of this material is permitted. Permission from IEEE must be obtained for all other uses, in any current or future media, including reprinting/republishing this material for advertising or promotional purposes, creating new collective works, for resale or redistribution to servers or lists, or reuse of any copyrighted component of this work in other works.}
\newcommand\copyrightnotice{%
	\begin{tikzpicture}[remember picture,overlay]
		\node[anchor=south,yshift=10pt] at (current page.south) {\fbox{\parbox{\dimexpr\textwidth-\fboxsep-\fboxrule\relax}{\copyrighttext}}};
	\end{tikzpicture}%
}

\begin{document}

\maketitle
\thispagestyle{empty}
\pagestyle{empty}

\AddToHookNext{shipout/background}{\copyrightnotice}

\begin{abstract}

One of the main limitations of multirotor UAVs is their short flight time due to battery constraints. A practical solution for continuous operation is to power the drone from the ground via a tether. While this approach has been demonstrated for stationary systems, scenarios with a fast-moving base vehicle or strong wind conditions require modeling the tether forces, including aerodynamic effects.
In this work, we propose two complementary approaches for low-latency quasi-static tether modeling with aerodynamics. The first is an analytical method based on catenary theory with a uniform drag assumption, achieving very fast solve times below 1~ms. The second is a numerical method that discretizes the tether into segments and lumped masses, solving the equilibrium equations using CasADi and IPOPT. By leveraging initialization strategies, such as warm starting and analytical initialization, low-latency performance was achieved with a solve time of 5~ms, while allowing for flexible force formulations.
Both approaches were validated in real-world tests using a load cell to measure the tether force. The results show that the analytical method provides sufficient accuracy for most tethered UAV applications with minimal computational cost, while the numerical method offers higher flexibility and physical accuracy when required. These approaches form a lightweight and extensible framework for low-latency tether simulation, applicable to both offline optimization and online tasks such as simulation, control, and trajectory planning.

\end{abstract}

\section{Introduction}
Multirotor drones are valuable tools in various fields, including inspection, mapping, and monitoring. However, in practice, they are limited to tasks that do not require continuous operation due to their limited flight time and battery constraints. Several approaches aim to solve this problem, for example, via autonomous landing and recharging. However, truly continuous operation can so far only be achieved by powering the drone from the ground via a tether. This method has been practically employed for many stationary base or slow-moving systems, but examples of fast-moving systems are less common \cite{marques_tethered_2023}. 

We aim to provide tools to optimize the design \cite{beffert_cable_2025}, modeling, and control of drones tethered to a moving ground vehicle in real-world conditions, such as strong wind. In these cases, it is crucial to know the forces from the tether acting on the drone due to tether weight, tension, and drag to ensure good system design, trajectory planning, and control. Therefore, the goal of this work is to create a simple and easily extendable low-latency tether simulation to estimate tether forces. The inputs of this simulation are the drone and ground vehicle position measured by RTK GPS, an air speed estimate, and the tether length, weight, diameter, and drag coefficient. Low-latency performance is important so that it can not only be used for offline optimization but also applications such as online wind estimation, model predictive control, trajectory planning, or SITL simulation.

\noindent The contributions of this work are:
\begin{itemize}
	\item A quasi-analytical tether model with 0.51~ms runtime
	\item A flexible quasi-static numerical solver with 5~ms runtime
	\item Experimental validation using onboard tether tension measurements
\end{itemize}

\section{Related Work}
Traditional applications of hanging wire modeling include the construction of power lines and bridges \cite{irvine_cable_1992}, where the focus is on horizontal cables subjected to vertical loads. While the basic mechanics are the same, tethered aerial vehicles involve primarily vertical cables with a combination of horizontal drag and gravity forces. A field that is much closer to UAVs and has seen a lot of research in this direction concerns kites and tethered airfoils for wind energy harvesting \cite{zanon_airborne_2014}, \cite{schmehl_analytical_2018}, \cite{koenemann_modeling_2017}. Knowing the tether forces is especially important in this field since these vehicles do not use active propulsion to stabilize the vehicle's position. Similarly, there is also some research specifically for modeling tethered multirotor UAVs \cite{dicembrini_modelling_2020}, \cite{borgese_tether-based_2022}, \cite{muttin_umbilical_2011}.

Classically, many different methods are used to solve the tether shape and forces \cite{fattori_tethered_2025}. Some are based on classical catenary theory, for example \cite{schmehl_analytical_2018}, that derives the quasi-analytical static solution for a tethered kite in a wind gradient. This approach has the benefit of providing a closed-form solution, but requires using the mean drag uniformly along the whole tether length, which is not physically accurate. With a solve time of approximately 1~s, this method is not feasible for low-latency applications. Another limitation is the lack of flexibility, as any change in the force equations would require redoing the mathematical derivation and because of the analytical nature only uniform forces can be modeled. Other examples include \cite{borgese_tether-based_2022} which measures the angle of the tether on both sides and uses a catenary model for localization and \cite{jain_tethered_2022} which models multiple drones connected by catenary tethers.

In \cite{koenemann_modeling_2017}, the authors model a quasi-static elastic tether as part of a landing trajectory optimal control problem. They represent the tether as lumped masses connected via elastic segments using per-segment cross-flow drag. A rigid-body velocity field is used to compute aerodynamic and centripetal loads along the tether. The tether model is integrated into a differential-algebraic equation (DAE) formulation and solved as part of the larger trajectory optimization using CasADi \cite{andersson_casadi_2012} and IPOPT \cite{wachter_implementation_2006}. This results in a physically accurate solution, but at increased computational cost. They note initialization challenges, which they solve by first solving a low-tension problem from a simple initial guess, then progressively increasing the target tension and solving again. They do not provide any runtime figures, but the iterative initialization approach, along with the problem size, makes it less feasible for low-latency use. In \cite{zanon_airborne_2014}, the authors use a similar approach to model and compare different configurations of tethered airfoils for wind energy generation.

In \cite{dicembrini_modelling_2020}, the tether is discretized into rigid segments and lumped masses, and the Lagrange equations are derived for dynamics, using cross-flow principles for lift and drag. They explicitly model the unwinding of the tether by allowing the first segment to change its length and dynamically adding or removing segments. The advantage of this method is that it provides physically accurate aerodynamics and dynamic cable behavior in 3D. However, it exhibits some numerical instability at higher wind speeds due to Euler angle singularities. Since dynamics are modeled, a small simulation time step of 0.1~ms is required. Due to the increased complexity of the problem, a segment length of 1.5~m was used to achieve a reasonable simulation speed; however, this limited the resolution of the results, suggesting that the method may not be suited for low-latency applications with suitable accuracy. Another limitation is that the complexity of the Lagrangian formulation limits flexibility when it comes to modifying forces or adding new ones. Similarly, \cite{muttin_umbilical_2011} uses discrete segments with lumped masses and finite element analysis to solve the dynamic tether shape for use in maritime oil spill detection.

As shown in Table~\ref{ComparisonTable}, analytical methods allow fast evaluation but rely on more assumptions, while quasi-analytical approaches have improved accuracy at similar computational cost. Numerical models with dynamics offer the highest physical fidelity but at significant computational cost, whereas quasi-static methods provide a balance between accuracy and low-latency performance. To the best of our knowledge, this is the first work to simultaneously provide per-segment aerodynamic drag modeling, millisecond-scale runtime suitable for low-latency use, and experimental validation using tension sensor measurements.

\begin{table*}[t]
	\centering
	\small
	\begin{tabular}{l l c c c c c c}
		\toprule
		Work & Method & Drag & Segment Drag & Dynamics & Flexibility & Low-Latency & Validation \\
		\midrule
		
		Borgese et al. \cite{borgese_tether-based_2022} 
		& analytical 
		& \textcolor{red}{no} 
		& \textcolor{red}{no} 
		& \textcolor{red}{no} 
		& \textcolor{red}{low} 
		& \textcolor{orange}{implied} 
		& \textcolor{green}{yes} \\
		
		Jain et al. \cite{jain_tethered_2022}
		& analytical 
		& \textcolor{red}{no} 
		& \textcolor{red}{no} 
		& \textcolor{red}{no} 
		& \textcolor{red}{low} 
		& \textcolor{red}{not stated} 
		& \textcolor{red}{no} \\
		
		Bigi et al. \cite{schmehl_analytical_2018}
		& quasi-analytical 
		& \textcolor{green}{yes} 
		& \textcolor{red}{no} 
		& \textcolor{red}{no} 
		& \textcolor{red}{low} 
		& \textcolor{red}{no (1~s)} 
		& \textcolor{red}{no} \\
		
		\textbf{Ours (analytical)} 
		& \textbf{quasi-analytical} 
		& \textbf{\textcolor{green}{yes}} 
		& \textbf{\textcolor{red}{no}} 
		& \textbf{\textcolor{red}{no}} 
		& \textbf{\textcolor{red}{low}} 
		& \textbf{\textcolor{darkgreen}{yes (0.51~ms)}} 
		& \textbf{\textcolor{green}{yes}} \\
		
		Koenemann et al. \cite{koenemann_modeling_2017}
		& numerical quasi-static 
		& \textcolor{green}{yes} 
		& \textcolor{green}{yes} 
		& \textcolor{red}{no} 
		& \textcolor{green}{good} 
		& \textcolor{red}{not stated} 
		& \textcolor{red}{no} \\
		
		Zanon et al. \cite{zanon_airborne_2014}
		& numerical quasi-static 
		& \textcolor{green}{yes} 
		& \textcolor{green}{yes} 
		& \textcolor{red}{no} 
		& \textcolor{green}{good} 
		& \textcolor{red}{not stated} 
		& \textcolor{red}{no} \\
		
		\textbf{Ours (numerical)} 
		& \textbf{numerical quasi-static} 
		& \textbf{\textcolor{green}{yes}} 
		& \textbf{\textcolor{green}{yes}} 
		& \textbf{\textcolor{red}{no}} 
		& \textbf{\textcolor{darkgreen}{excellent}} 
		& \textbf{\textcolor{green}{yes (5.4~ms)}} 
		& \textbf{\textcolor{green}{yes}} \\
		
		Dicembrini et al. \cite{dicembrini_modelling_2020}
		& numerical dynamic 
		& \textcolor{green}{yes} 
		& \textcolor{green}{yes} 
		& \textcolor{green}{yes} 
		& \textcolor{green}{good} 
		& \textcolor{orange}{unclear} 
		& \textcolor{red}{no} \\
		
		Muttini et al. \cite{muttin_umbilical_2011}
		& numerical dynamic 
		& \textcolor{green}{yes} 
		& \textcolor{green}{yes} 
		& \textcolor{green}{yes} 
		& \textcolor{green}{good} 
		& \textcolor{orange}{with low fidelity} 
		& \textcolor{red}{no} \\
		
		\bottomrule
	\end{tabular}
	\caption{Comparison of tether modeling approaches. Modeling of drag, per-segment drag and dynamics indicate best physical accuracy.}
	\label{ComparisonTable}
\end{table*}

\section{Methods}
The shape and tension of a hanging wire are the result of the differential force equations at every point along the cable. These forces include the weight of the cable, the tension in the cable, and any additional forces like drag or lift. In the case where there is a uniform mass distribution along the cable and only gravity acts on it, the tether follows the catenary shape given by
\begin{equation}\label{catenary}
	a*\cosh((x-x_0)/a)+y_0
\end{equation}
\begin{equation}
	a=H/w
\end{equation}
\begin{itemize}
	\item $(x_0,\, y_0)$ is the offset of the catenary
	\item $(x_0,\, y_0 + a)$ is the lowest point of the catenary (vertex)
	\item $a>0$ is the shape parameter
	\item $H$ is the constant horizontal tension
	\item $w$ is the cable weight per unit length
\end{itemize}

\subsection{Analytical Approach}
This function is used to obtain an analytical solution by determining the parameters $a, (x_0,\, y_0)$ that satisfy the constraints.
We define point $p_1=(x_1,\, y_1)$ as the position of the ground vehicle and $p_2=(x_2,\, y_2)$ as the position of the drone. For simplicity, we will furthermore assume that $y_1<y_2$. Both $p_1$ and $p_2$ must lie on the catenary curve \eqref{catenary} resulting in constraints \eqref{c1} and \eqref{c2}.
The arc length between $p_1$ and $p_2$ must be equal to the wire length $L$ resulting in constraint \eqref{c3}. By solving the nonlinear system of equations, the parameters that result in a valid catenary are determined.
\begin{align}
	&a * \cosh((x_1 - x_0) / a) + y_0 - y_1 = 0\label{c1}\\
	&a * \cosh((x_2 - x_0) / a) + y_0 - y_2 = 0\label{c2}\\
	&a * |\sinh((x_2 - x_0) / a) - \sinh((x_1 - x_0) / a)| - L = 0\label{c3}
\end{align}

Due to the low computational complexity of solving the problem, very short solve times below 1~ms (Fig.~\ref{TimeComparison}) were achieved by using a lightweight solver ({\it scipy.fsolve} a Newton-type root finding method). It is common for numerical methods to vary greatly depending on the quality of the initialization or initial guess. In this case, the solver can sometimes struggle to find a solution, especially if the problem becomes ill-posed (when the start and end points are close horizontally). To mitigate this, multiple parameter sets are used as fallback for the initial guess in case the solve fails (Table~\ref{AnalyticalGuessFormulas}). Because of the sub-millisecond solve time and early termination in case of bad initialization, attempting multiple solves does not have a significant impact on the runtime.

\begin{table}[!h]
	\centering
	\begin{tabular}{ll}
		\toprule
		\textbf{Type} & \textbf{Parameter Guess} ($a$, $x_0$, $y_0$) \\
		\midrule
		Parabolic 1x & $(a_{par},\, x_{mid},\, y_1-s_{par}-a_{par})$ \\
		Parabolic Nx & $(N*a_{par},\, x_{mid},\, y_1-s_{par}-N*a_{par})$ \\
		No Sag 1x & $(a_{nosag},\, x_{mid},\, y_1-a_{nosag})$ \\
		No Sag Nx & $(N*a_{nosag},\, x_{mid},\, y_1-N*a_{nosag})$ \\
		Large Sag & $(L/8,\, x_{mid},\, y_1-L/2)$ \\
		\bottomrule
	\end{tabular}
	\caption{Initial guesses for the analytical method based on parabolic approximation and heuristics.}
	\label{AnalyticalGuessFormulas}
\end{table}

Generally $x_0$ is guessed to be at the midpoint $x_{mid}=(x_1+x_2)/2$ between $x_1$ and $x_2$. The shape parameter $a$ is guessed using \eqref{apar}, which is a parabolic approximation of the catenary based on the second-order Taylor expansion (\hyperref[derivation]{Appendix~A}). The tether sag is estimated by \eqref{asag}. 

\noindent Then $y_0=y_1-s-a$ is estimated following \eqref{catenary} since the sag is the vertical distance from $y_1$ to the vertex and $y_0$ is $a$ below that. To cover different cases, an option that assumes zero sag $a_{nosag}$ as well as different scaling factors for $a$ are used. There is also a heuristic guess for large sag that does not use the parabolic approximation, instead assuming that the sag is approximately half the cable length.

\begin{equation}\label{apar}
	a_{par}=\frac{d_x^2}{8*s_{par}}
\end{equation}
\begin{equation}\label{asag}
	s_{par}=\sqrt{\frac{3}{8}*d_x*(L-d_x)}
\end{equation}
\begin{itemize}
	\item $d_x=(x_2-x_1)$ is the horizontal distance between $p_1$ and $p_2$
	\item $s_{par}$ is the estimated sag based on the parabolic cable length equation $L = d_x + (8s^2)/(3d_x)$ \cite{irvine_cable_1992}
\end{itemize}

The different initial guesses from Table~\ref{AnalyticalGuessFormulas} were compared on a real-world flight (see Sec. \ref{Experiments}) to determine the number of cases in which the guess produced the fastest solve time. This was done to determine the order in which the guesses should be attempted to speed up the solve by not wasting time on bad guesses. The results are listed in Table~\ref{AnalyticalGuessTimes}.

\begin{table}[htbp]
	\centering
	\begin{tabular}{lc}
		\toprule
		Method & Percentage (\%) \\
		\midrule
		Parabolic 2x & 42.9 \% \\
		Parabolic 1.2x & 27.1 \% \\
		Large Sag & 10.0 \% \\
		No Sag 1x & 7.1 \% \\
		No Sag 2x & 4.7 \% \\
		Parabolic 0.8x & 4.7 \% \\
		Parabolic 1x & 3.5 \% \\
		\bottomrule
	\end{tabular}
	\caption{Comparison between the initial guess strategies for the analytical method from Table~\ref{AnalyticalGuessFormulas}. Shown is the percentage of cases in which the method produced the fastest solve time. The test was run on a real-world flight to represent the actual distribution of cases.}
	\label{AnalyticalGuessTimes}
\end{table}

The current catenary formulation does not model any drag; however, in the case of real-world tethered drones, having zero airspeed is uncommon due to wind and the drone's movement. Drag can be added to the catenary model by making the simplification that the aerodynamic force acts on the whole tether equally. This means that the drag behaves like a potential field that can be combined with the potential field of the tether weight. By doing this, the resulting shape is still a catenary curve, but in a rotated coordinate system where the combined potential field points down. So by first rotating the start and end points and applying the combined force instead of just the weight, the same catenary equation can be used. Points or tension values can then be sampled in the rotated coordinate system and rotated back to produce the simulation result.

The total drag on the tether is calculated using the Rayleigh drag equation~\eqref{DragEquation} with the projected vertical tether length as an approximation of the exposed cross-flow area.

\begin{equation}\label{DragEquation}
	F_D=\frac{1}{2}*\rho*C_D*A*V^2
\end{equation}

This approach underestimates the drag in cases where the tether sags below the start point. To obtain an accurate solution, an initial solve is performed using the vertical distance between $p_1$ and $p_2$ as an estimate for the vertical length of the tether. The actual vertical length is then computed numerically and used for the next iteration until the error between the estimated and actual vertical length used to calculate the drag converges. This ensures that the total amount of drag on the tether is correct.

In conclusion, the analytical model makes the assumptions of an inextensible cable without stretch, uniform drag, and a uniform mass distribution along the tether.

\subsection{Numerical Approach}
In reality, the drag on the cable is not uniform. To achieve a physically accurate solution, it is necessary to include varying forces along the tether length. The analytical approach does not allow this since the resulting shape will not be a catenary. Therefore, we use CasADi \cite{andersson_casadi_2012}, which allows defining nonlinear optimization problems symbolically and uses IPOPT \cite{wachter_implementation_2006} to solve them. For this numerical approach, the cable is split into discrete point masses connected via rigid segments. Since quasi-static equilibrium with uniform airflow direction is assumed, the tether will remain in a single vertical plane, allowing for a 2D formulation.

One way to define the problem would be to minimize the energy in the system while maintaining the length and endpoints. The downside of this approach is that it makes it harder to implement arbitrary forces, as they would have to be converted into energy, which would depend on the deflection from the rest state. Instead, a more practical approach is to redefine the problem in terms of constraint satisfaction, without requiring minimization. In this case, the positions and tension are solved directly so that the net force at each node is zero. Since an inextensible tether is assumed, the tension forces can be solved directly, in contrast to classical spring mass models. However, without a spring force to balance segment lengths, an explicit segment length constraint is necessary. By limiting the tension to positive values, the numerical stability is improved, and physically accurate solutions are ensured. This is necessary, as cables can only act in tension, not compression. 

First, the segment forces are calculated by combining the drag and tension from the solver. The drag in this case is still based on the Rayleigh drag equation~\eqref{DragEquation}, but accurately uses the vertical length of each segment to calculate the exposed area. 
\begin{equation}
	A_{exp}=d*\Delta y
\end{equation}
\begin{itemize}
	\item $A_{exp}$ is the area exposed to the airflow
	\item $d$ is the tether diameter
	\item $\Delta y$ is the segment vertical component (assuming horizontal wind)
\end{itemize}

At this point, other segment forces like lift or drag based on a wind gradient could be added. The forces on each node are then summed up with a half contribution from each connected segment, plus gravity and potentially any other node-based forces like inertia. A constraint is then generated for each node to achieve zero net force.

These constraints are implemented symbolically using CasADi, which constructs the Lagrangian, applies automatic differentiation to compute the objective function gradient, constraint Jacobian, and Lagrangian Hessian, then solves the optimization problem using IPOPT. The objective function is defined as a constant zero, reducing the problem to pure constraint satisfaction. This formulation is equivalent to solving a nonlinear system but allows leveraging IPOPT’s robust constraint handling and automatic differentiation. The key to achieving fast solve times is to cache and reuse the CasADi solver, so the problem setup and automatic differentiation do not have to be repeated and only need to be numerically evaluated for each solve.

\begin{figure}[htbp]
	\centering
	\includegraphics[width=0.3\textwidth]{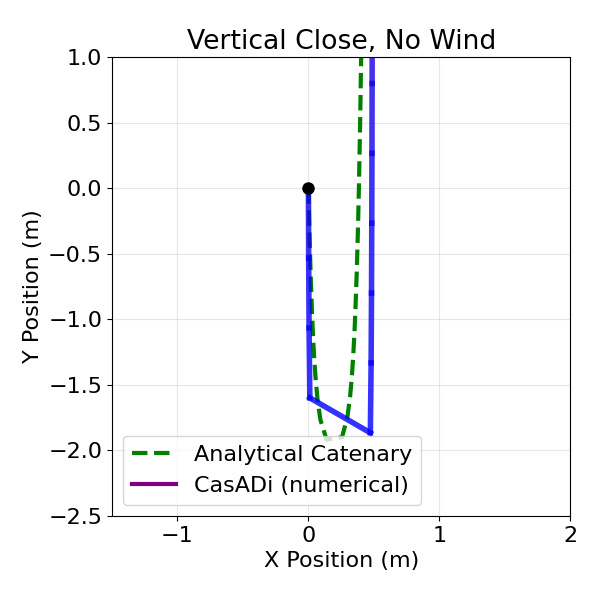}
	\caption{Example where the start and end points are almost vertical and there is no wind. The fixed segment length of the discrete approach limits the minimum bend radius.}
	\label{Segment}
\end{figure}

One inherent limitation of any discrete approach is that the segment length limits the maximum bend radius (Fig.~\ref{Segment}). This becomes apparent when the airspeed is low and the start and end are close together horizontally, resulting in a sharp bend at the bottom. The numerical approach will still produce a result, but the forces will not be physically accurate. Therefore, it is better to use the analytical method if the horizontal distance between the start and end points is shorter than twice the segment length. By doing this check in the rotated coordinate system (as described in the analytical approach), cases with low sag and therefore without a sharp bend are handled correctly.

As previously mentioned, initialization has a significant impact on numerical methods; the same is true for this approach. Unlike the analytical method, it converges even with bad initialization, but the solve time can vary significantly. To make sure the best possible initialization is used, there are two options: using the analytical solution for initialization and using the solution from the previous time step (warm start). When using the analytical solution, it is crucial to sample the points at equal arc lengths instead of just uniformly along the x-axis. The same applies to calculating the tensions where the segment midpoints need to be sampled.

Compared to the analytical method, fewer assumptions need to be made since any arbitrary force could be added according to the decision diagram in (\hyperref[derivation]{Appendix~C}), to make the simulation more accurate. However, it has to be tested if the solver is stable for a specific arbitrary force. Similarly, by removing the length constraints and adding a spring force, a flexible tether could be modeled.

For our implementation, we used the planar case where the drone, ground vehicle and airspeed direction lie in a 2D plane, as this corresponds with our validation test. In general this assumption is not necessary and the problem can be formulated in 3D as well by adding another dimension to the forces and positions.

\section{Experiments and Results}
\label{Experiments}

\begin{figure*}[!t]
	\centering
	\begin{minipage}{0.82\textwidth}
		\begin{subfigure}[b]{0.32\textwidth}
			\includegraphics[width=\textwidth]{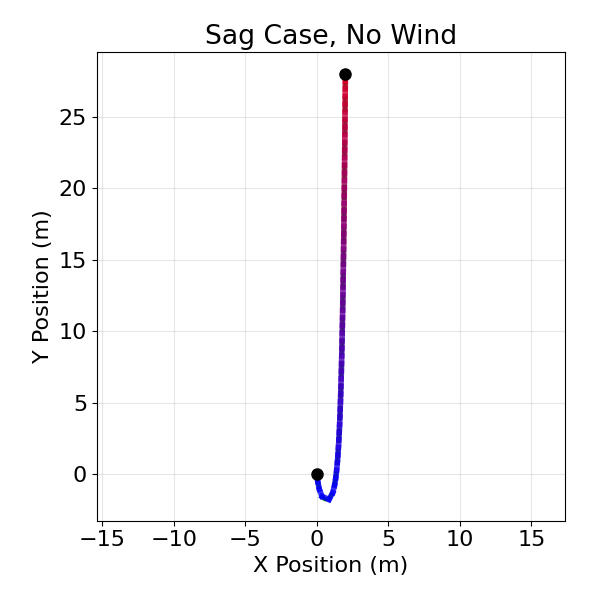}
			\caption{}
			\label{example_a}
		\end{subfigure}
		\hfill
		\begin{subfigure}[b]{0.32\textwidth}
			\includegraphics[width=\textwidth]{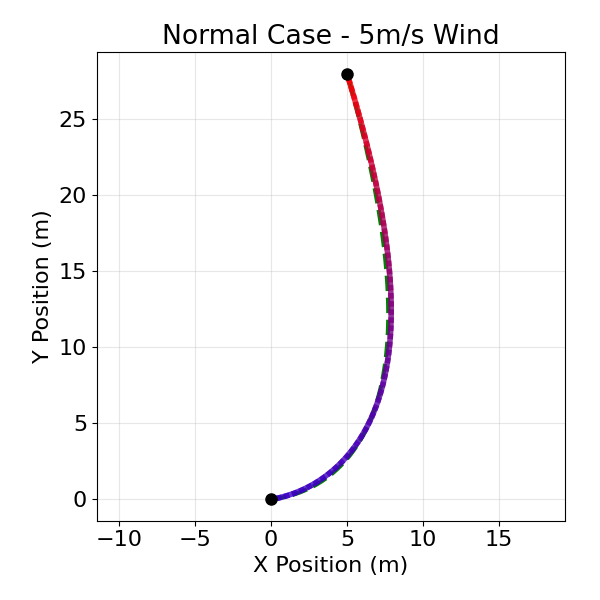}
			\caption{}
			\label{example_b}
		\end{subfigure}
		\hfill
		\begin{subfigure}[b]{0.32\textwidth}
			\includegraphics[width=\textwidth]{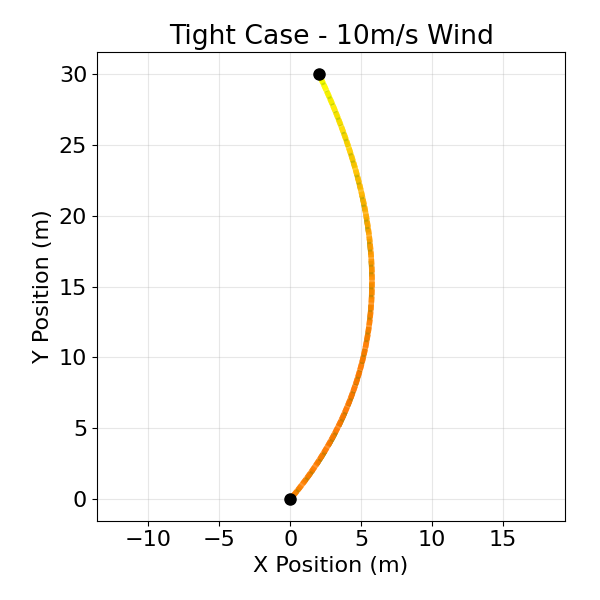}
			\caption{}
			\label{example_c}
		\end{subfigure}
		
		\vspace{0.5em}
		
		\begin{subfigure}[b]{0.32\textwidth}
			\includegraphics[width=\textwidth]{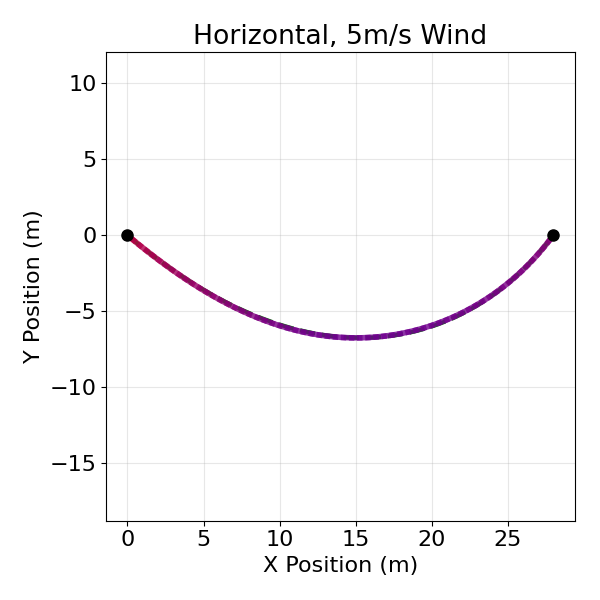}
			\caption{}
			\label{example_d}
		\end{subfigure}
		\hfill
		\begin{subfigure}[b]{0.32\textwidth}
			\includegraphics[width=\textwidth]{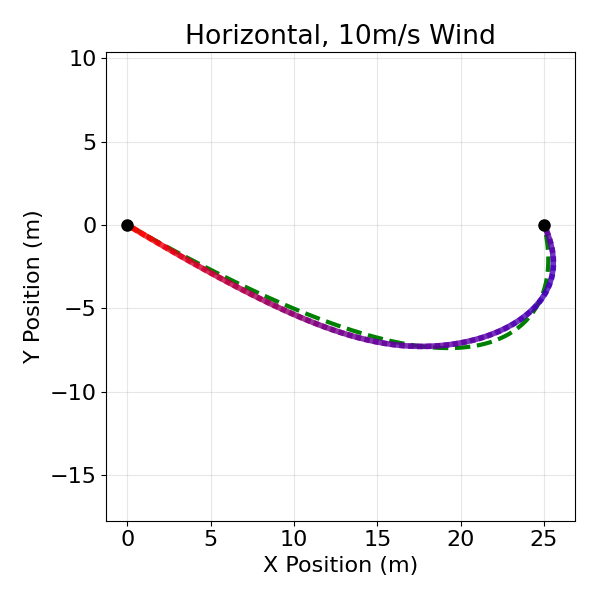}
			\caption{}
			\label{example_e}
		\end{subfigure}
		\hfill
		\begin{subfigure}[b]{0.32\textwidth}
			\includegraphics[width=\textwidth]{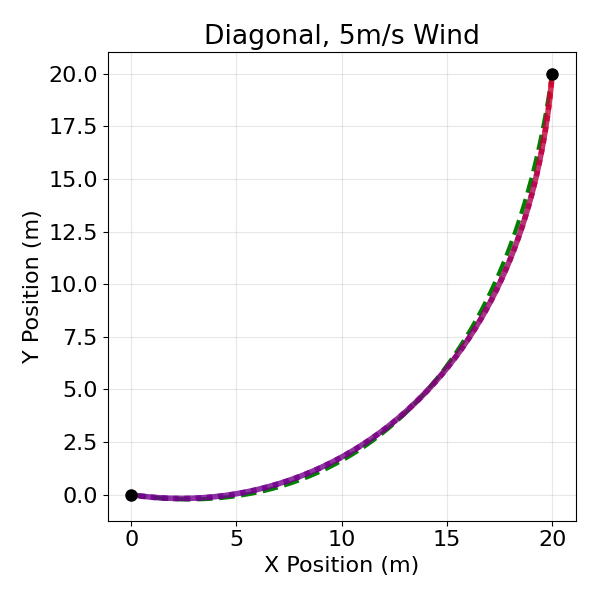}
			\caption{}
			\label{example_f}
		\end{subfigure}
	\end{minipage}
	\hfill
	\begin{minipage}[c]{0.16\textwidth}
		\centering
		\includegraphics[width=\textwidth]{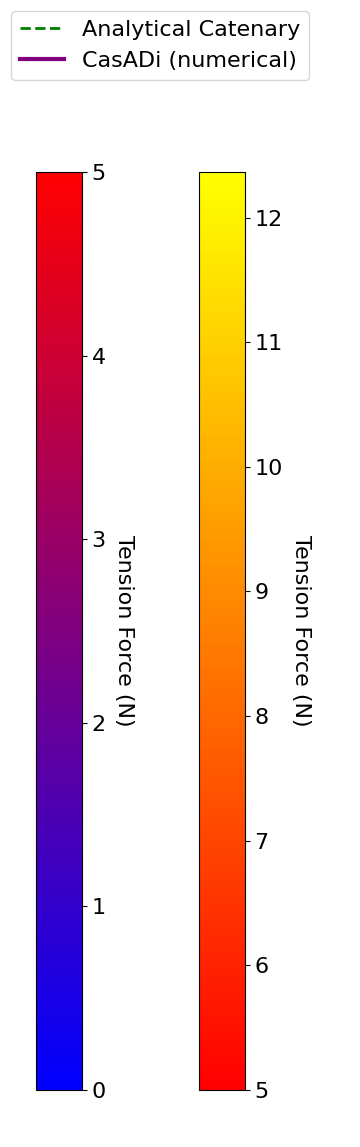}
	\end{minipage}
	
	\caption{Examples of the shape and tension of different configurations solved via the analytical and numerical (CasADi) approach. \ref{example_a} and \ref{example_b} show typical cable behaviour for a tethered UAV, while \ref{example_c} highlights the high forces on tight cables. The comparison between \ref{example_d} and \ref{example_e}, \ref{example_f} highlight that the analytical solution is less accurate for large horizontal distances and strong winds. Note that the color bars are scaled differently.}
	\label{Examples}
\end{figure*}

Figure~\ref{Examples} shows the tether shape and tension along the tether for different configurations. It also highlights that the analytical solution is very close to the numerical one in most cases. Another interesting insight from the plots is that the forces can exceed the tether weight by a large margin if there is strong wind and the tether gets tight (\ref{example_c}). This case could be detected on the ground by measuring the angle of the tether; the steeper it is, the higher the expected tension. In this case, the forces should be lowered by decreasing the drone's altitude. The comparison between \ref{example_d} and \ref{example_e}, and \ref{example_f} highlights that there are some differences between the analytical and numerical solution in cases with large horizontal distance and strong winds. This is due to the uniform drag assumption of the analytical approach, which becomes apparent when there are more almost horizontal segments.

\begin{figure}[htbp]
	\centering
	\includegraphics[width=0.5\textwidth]{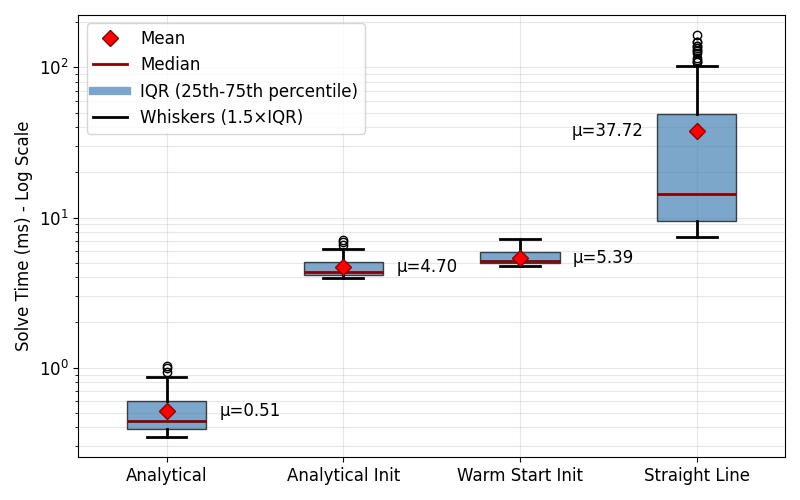}
	\caption{Comparison of solve times for the analytical approach, numerical approach with analytical initialization, numerical approach with warm start, and the numerical approach initialized with a straight line. The time to calculate the numerical initialization is not included. Note that the y-axis is on a logarithmic scale. The numerical method was run for 60 segments. It shows that the numerical approach is roughly 10 times slower than the analytical one, and using unsuitable initialization can further increase the solve time by a factor of 8.}
	\label{TimeComparison}
\end{figure}

Figure \ref{TimeComparison} has the solve times for a real-world flight using our python implementation running single-threaded on an i7-9700 CPU. It shows that the analytical method is very fast, with a mean runtime of 0.51~ms, even with the iterative approach and using multiple guesses. This makes it feasible to use the analytical solution to initialize the numerical solver without a significant impact on the solve time. In this case, the mean is 4.70~ms (in addition to the time necessary to calculate the analytical initialization). In case the solution from the previous time-step is used to initialize the numerical approach, the runtime is similarly fast at a mean of 5.39~ms. In contrast, using a bad initialization, for example, just a straight line between start and end point, has a significant impact on the solve time, with a mean of 37.72~ms and a wide distribution of values. In summary, the numerical method is about 10 times slower than the analytical one, and improper initialization can further increase the numerical solve time by a factor of 8.

Besides setting {\it ipopt.tol} and {\it ipopt.constr\_viol\_tol} to $1\times\nobreak 10^{-8}$ to match the desired residual threshold of the analytical method, default solver settings were used. Further improvements of the numerical solve time may be possible through IPOPT parameter tuning.

\begin{figure}[htbp]
	\centering
	\includegraphics[width=0.37\textwidth]{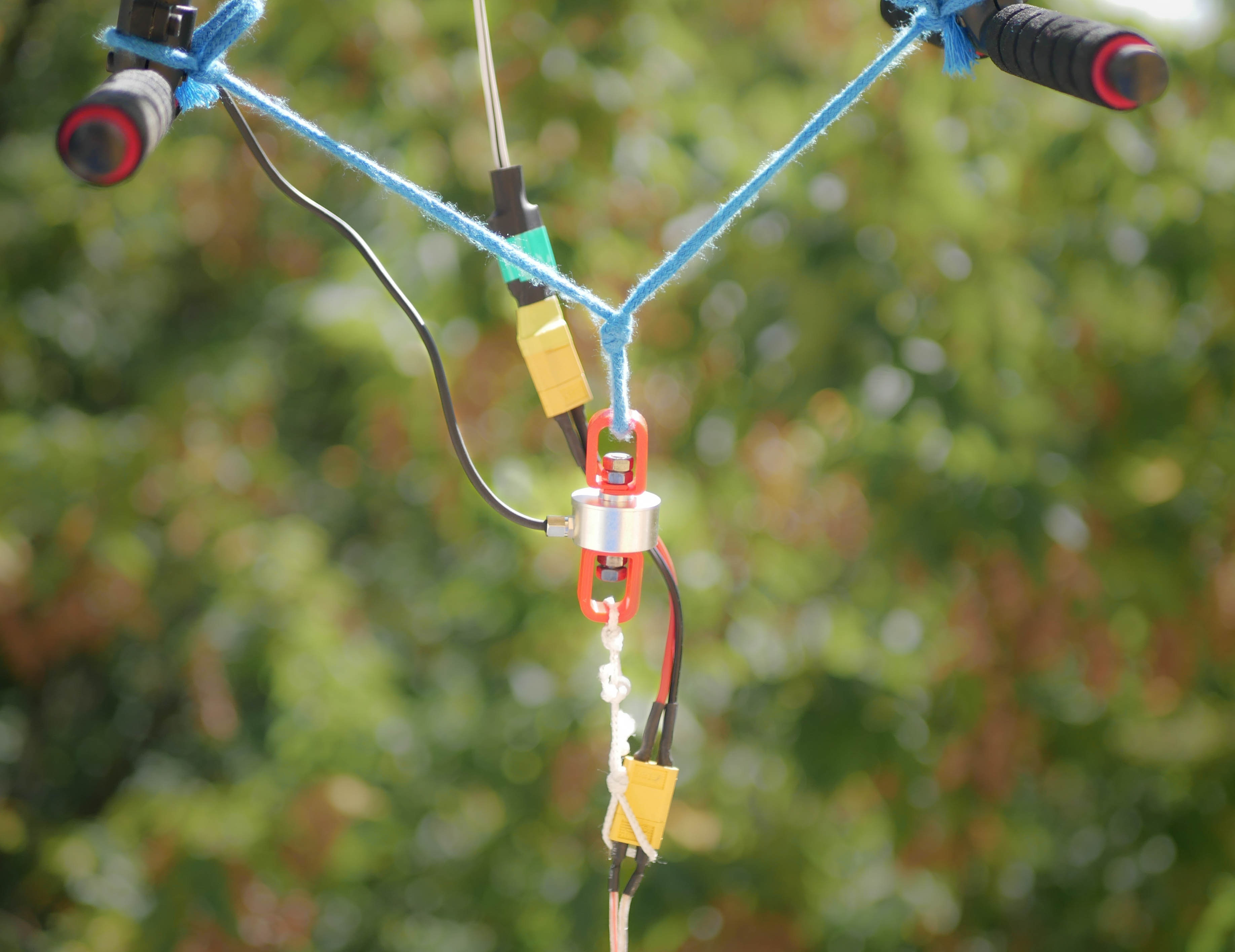}
	\caption{Photo of the load cell sensor attached to the drone for measuring the tether tension.}
	\label{picture}
\end{figure}

To confirm the simulation results, a sensor was built to measure the tension in the cable on the drone side. It uses an inline load cell rated for 10~kg, a NAU7802 analog-digital converter, and an RP2040 microcontroller (Fig.~\ref{picture}). The tension is measured at a rate of 10~Hz and sent to the drone flight controller via MAVLink, where it is logged.

\begin{figure}[htbp]
	\centering
	\includegraphics[width=0.5\textwidth]{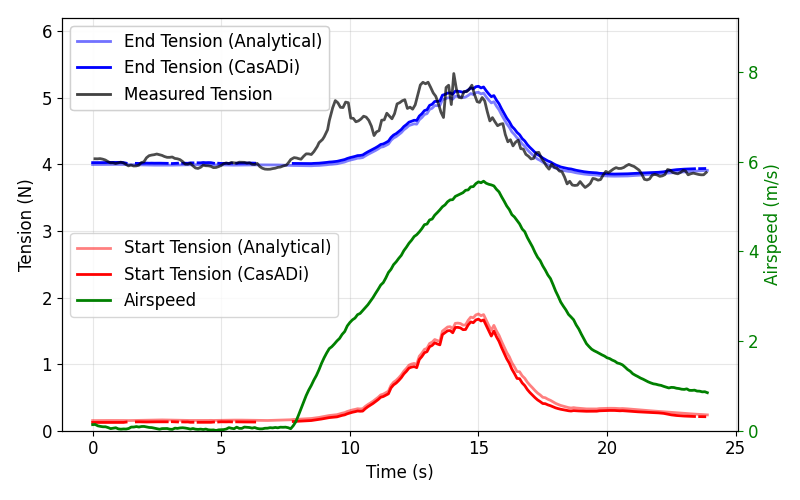}
	\caption{Comparison between the tension from the analytical and numerical (CasADi) method on RTK GPS data from a real flight. Airspeed is shown in green. Start force (red) on the ground and end force (blue) on the drone demonstrate close agreement between the analytical and numerical method. The tension was measured on the drone (black) and validates the simulation but shows some discrepancies during acceleration due to unmodeled system dynamics.}
	\label{TensionComparison}
\end{figure}

A test flight was conducted with an aluminum cable of length 32~m and mass 430~g at a flight height of 30~m. This was done during low wind conditions, so the airspeed could be approximated by drone velocity. Low wind conditions were only required to simplify the experimental verification. For the simulation, the source of the relative airspeed (drone motion, wind, or a combination of both) is irrelevant. The results of this test flight are shown in Fig.~\ref{TensionComparison} and demonstrate close agreement between the analytical and numerical solutions. The start force has a maximum absolute error of 0.111~N with an RMSE of 0.049~N, while the end force exhibits a maximum absolute error of 0.093~N with an RMSE of 0.048~N. For the end tension, the NRMSE (normalized by the maximum measured tension) is below 1.25\%. This confirms that the analytical solution is sufficiently accurate when the ratio between vertical and horizontal distance is high, as in the case for tethered drones (see \hyperref[ErrorExploration]{Appendix~B} for a systematic error exploration). The analytical solution is therefore sufficient when minimal solve time is required or computational resources are limited.

To match the simulation to the real-world measurements, system identification was done to estimate the drag coefficient of the cable. The optimization was carried out using {\it scipy.minimize} with the L-BFGS-B algorithm. It is a quasi-Newton method well suited for problems with computationally expensive cost functions. This further illustrates the benefit of a fast simulation framework, since evaluating the loss function requires running the simulation for all data points. A total of 240 data points were used with an average time of 1.5~s per cost function evaluation. Convergence was achieved after 4 iterations with a total optimization time of 12~s. The identified drag coefficient was 1.16, which lies within the expected range.

\begin{figure}[htbp]
	\centering
	\includegraphics[width=0.5\textwidth]{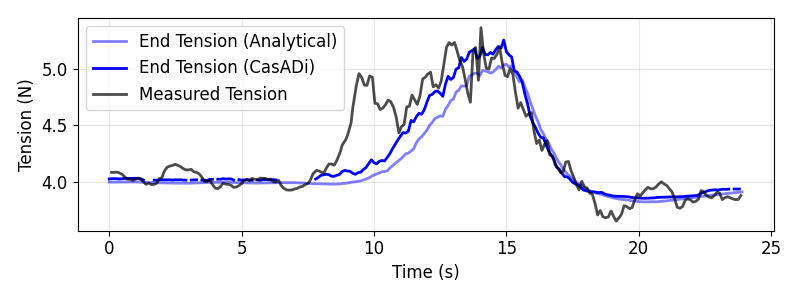}
	\caption{Adding a simplified inertial term to the numerical (CasADi) solution shows improved physical accuracy in transient cases compared to the analytical solution without dynamics. The remaining discrepancies are hypothesized to be caused by whipping of the cable which is not modeled.}
	\label{TensionComparison_inertia}
\end{figure}

The load cell measurements are somewhat noisy but confirm the simulation results. In addition to measurement noise, deviations can be attributed to unexpected wind and GPS altitude errors. When the tether is almost tight, small changes in altitude can strongly affect the tension, so any mismatch between measured and actual altitude will impact the tension accuracy of the simulation.
We can also observe a discrepancy due to the unmodeled dynamic behavior of the cable, which causes more tension during acceleration and less during deceleration. This is expected and would be less apparent during extended flight at constant velocity.
\newpage
To demonstrate the flexibility of the approach and as an attempt to model some of the system dynamics, we implemented a simplified inertia force. Instead of calculating the velocities and accelerations of individual nodes, we use the acceleration of the drone to calculate the inertial forces for the whole tether. The only implementation change necessary to accomplish this was passing the acceleration into the solver and using it to calculate the inertial force from the node mass. This was achieved with minimal changes to 4 lines of the implementation, showing the flexibility of the approach. System identification was performed again to identify a corrected drag coefficient of 1.13. Indeed, the results with the added dynamics more closely match the measured tension (Fig.~\ref{TensionComparison_inertia}). During acceleration we observe a period of increased tension that is not present in the simulation results. We hypothesize that this effect is caused by whipping of the cable, producing short-term localized tension peaks before the forces redistribute along the cable. By transitioning from the simplified dynamics to a full per-node dynamic model, this effect could be captured.

\newpage
\section{Conclusion}
In this work, two complementary approaches for quasi-static low-latency tether simulation were proposed. The first is an analytical method based on catenary theory with a uniform drag approximation, achieving 0.51~ms solve times while maintaining physical accuracy for typical tethered UAV applications. For these cases, the analytical solution exhibits a NRMSE of 1.25\% compared to the more physically accurate numerical method, making it suitable for low-latency applications with limited computational resources.

A numerical approach was also introduced to provide increased physical fidelity. Using CasADi to solve a discretized quasi-static formulation with lumped masses and rigid segments, solve times of 5~ms were achieved by leveraging warm starting and analytical initialization. An additional strength of this method is its flexibility in allowing easy modification and extension of force equations.

Experimental validation using load cell measurements on the drone confirmed the accuracy of both approaches. Deviations observed during transient maneuvers are primarily attributable to the lack of dynamic modeling. Adding a simplified inertial term reduced these discrepancies and demonstrated the extensibility of the numerical framework.

The proposed method assumes quasi-static equilibrium and does not capture high-frequency cable dynamics such as whipping or out-of-plane 3D motion. Although the simplified inertial extension improves transient behavior, fully dynamic modeling would be required for aggressive maneuvers or highly variable wind conditions. Extending the numerical framework in this regard while maintaining low-latency capability is a direction for future work. Besides improving the method, future work includes integration into online wind estimation, model predictive control, or SITL simulation environments.

As model complexity increases, real-world validation using a single tension measurement becomes less indicative of overall model accuracy. More comprehensive validation strategies should therefore be examined in future work such as distributed tension measurements or shape reconstruction via vision.

Overall, the proposed methods provide a physically consistent and extensible framework for low-latency tether simulation, validated through experimental tension measurements, making it suitable for both offline analysis and control applications in tethered UAV systems.

\newpage
\bibliographystyle{IEEEtran}
\bibliography{references}

\newpage
\section*{APPENDIX}
\subsection{Derivation of \eqref{asag}:}\label{derivation}
\begin{align*}
\cosh x&=1+\frac{x^2}{2!}+\frac{x^4}{4!}+\cdots\\
y(x)&=a*\cosh\frac{x}{a}\\
&\approx a*(1+\frac{(x/a)^2}{2})\\
&= a+\frac{x^2}{2a}
\end{align*}
This is the parabolic approximation around the vertex. For a symmetric catenary the sag $s$ is the distance between the vertex at $y_0+a=a$ and start point at $y(d_x/2)$.
\begin{align*}
	s&=y(d_x/2)-a\\
	&\approx \frac{(d_x/2)^2}{2a}\\
	a&\approx \frac{(d_x/2)^2}{2s}=\frac{d_x^2}{8s}
\end{align*}
\\

\subsection{Analytical Error Exploration}
\begin{figure}[htbp]
	\centering
	\includegraphics[width=0.5\textwidth]{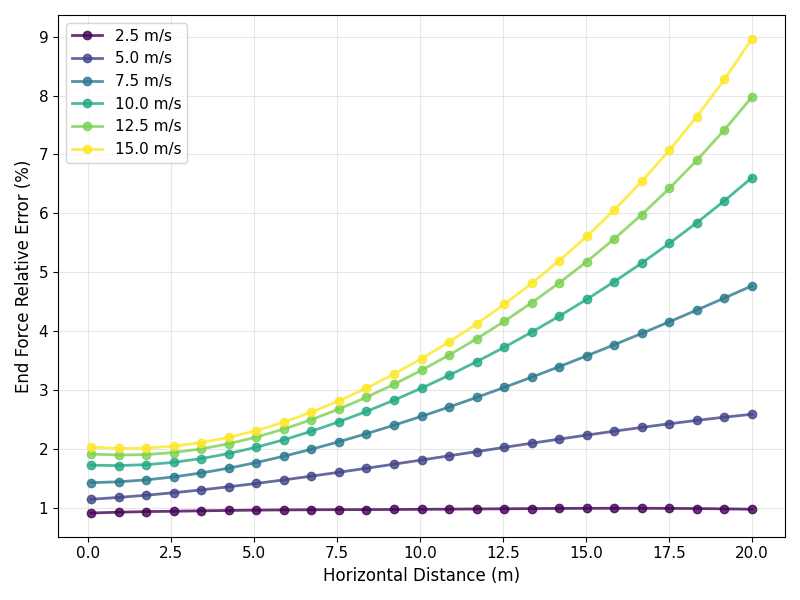}
	\caption{The relative end force error of the analytical solution compared to the numerical solution. It was calculated for a 32~m cable with a total distance between the drone and ground vehicle of 30~m. It shows that the error increases with airspeed and horizontal distance. This demonstrates that for common tethered drone applications with low horizontal distances the analytical method produces accurate results.}
	\label{ErrorExploration}
\end{figure}

\newpage
\subsection{Decision Diagram}
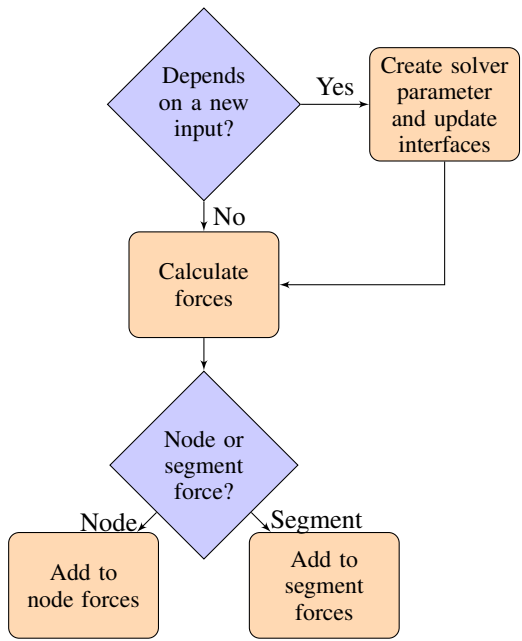
\begin{figure}[htbp]
	\centering
	\begin{tikzpicture}[node distance = 2cm, auto, scale=0.8]
		\node [decision] (input) at (0,0) {Depends\\ on a new input?};
		\node [process] (create) at (4,0) {Create solver parameter and update interfaces};
		\node [process] (calculate) at (0,-3.3) {Calculate forces};
		\node [decision] (type) at (0,-6.5) {Node or segment force?};
		\node [process] (node) at (-2,-9) {Add to node forces};
		\node [process] (segment) at (2,-9) {Add to segment forces};
		
		\path [line] (input) -- node[above] {Yes} (create);
		\path [line] (input) -- node[right] {No} (calculate);
		\path [line] (create) |- (calculate);
		\path [line] (calculate) -- (type);
		\path [line] (type) -- node[left] {Node} (node);
		\path [line] (type) -- node[right] {Segment} (segment);
		
	\end{tikzpicture}
	\caption{Decision diagram when adding new forces to the simulation. It shows the flexibility of the method, allowing for easy modifications.}
	\label{DecisionDiagram}
\end{figure}

\end{document}